\documentclass[conference]{IEEEtran}
\IEEEoverridecommandlockouts
\usepackage{cite}
\usepackage{amsmath,amssymb,amsfonts}
\usepackage{algorithmic}
\usepackage{graphicx}
\usepackage{textcomp}
\usepackage{multirow}
\usepackage{booktabs}
\usepackage{rotating}
\usepackage{tablefootnote}
\usepackage[hidelinks]{hyperref}

\begin{document}

\title{D-CTNet: A Dual-Branch Channel-Temporal Forecasting Network with Frequency-Domain Correction}
\author{
    \IEEEauthorblockN{
        Shaoxun Wang\textsuperscript{\dag*}, 
        Xingjun Zhang\textsuperscript{\dag*}, 
        Kun Xia\textsuperscript{\dag}, 
        Qianyang Li\textsuperscript{\dag}, 
        Jiawei Cao\textsuperscript{\dag}, and 
        Zhendong Tan\textsuperscript{\dag}
    }
    \IEEEauthorblockA{
        \textsuperscript{\dag}School of Computer Science and Technology, Xi'an Jiaotong University, Xi'an, China \\
        \{shaoxunwang, liqianyang, 3124351080, 772316639\}@stu.xjtu.edu.cn, \{xjzhang, xiakun\}@xjtu.edu.cn
    }
}

\maketitle

\begin{abstract}
Accurate Multivariate Time Series (MTS) forecasting is crucial for collaborative design of complex systems, Digital Twin building, and maintenance ahead of time. However, the collaborative industrial environment presents new challenges for MTS forecasting models: models should decouple complex inter-variable dependencies while addressing non-stationary distribution shift brought by environmental changes. To address these challenges and improve collaborative sensing reliability, we propose a Patch-Based Dual-Branch Channel-Temporal Forecasting Network (D-CTNet). Particularly, with a parallel dual-branch design incorporating linear temporal modeling layer and channel attention mechanism, our method explicitly decouples and jointly learns intra-channel temporal evolution patterns and dynamic multivariate correlations. Furthermore, a global patch attention fusion module goes beyond the local window scope to model long range dependencies. Most importantly, aiming at non-stationarity, a Frequency-Domain Stationarity Correction mechanism adaptively suppresses distribution shift impacts from environment change by spectrum alignment. Evaluations on seven benchmark datasets show that our model achieves better forecasting accuracy and robustness compared with state-of-the-art methods. Our work shows great promise as a new forecasting engine for industrial collaborative systems. Code is available at \href{https://github.com/shaoxun6033/DCTNet}{https://github.com/shaoxun6033/DCTNet}.

\end{abstract}


\section{Introduction}
\label{sec:intro}

With the emergence of Industry 4.0 and Cyber-Physical Systems \cite{oks2024cyber}, industrial design and manufacturing are transitioning from fixed and static collaboration to dynamic cooperation. For complicated collaborative scenarios, including smart grid management \cite{rathor2020energy}, manufacturing pipeline monitoring \cite{mahmoud2025comprehensive, bhaskaran2020future}, and supply chain coordination \cite{sehrawat2024predicting, ghodake2024enhancing}, efficient system coordination and operation \cite{wang2025survey,li2025dpanet} highly depend on massive streams of data collected from large-scale sensor networks to enable collaborative cooperation among various subsystems. As the main carrier of massive streams of data, MTS contains valuable state information of system operation. Therefore, accurate and reliable time series forecasting is no longer an ordinary data science problem, but an enabling technology to support Digital Twin construction \cite{luo2025applications, moshood2024infrastructure}, collaborative decision making in IIoT \cite{channi2024deep} and predictive maintenance of complex systems \cite{benhanifia2025systematic}. 

However, accurately forecasting MTS in various practical collaborative design and industrial applications faces tremendous challenges. First, industrial systems usually involve multiple coupled subsystems, making dynamic dependencies in variable dimension (channel dimension) complicated \cite{yi2024deep}. Second, due to the openness of collaborative systems, dynamic changes in data distribution also occur in temporal dimension \cite{ma2024u, fan2025flow}. Although certain deep learning methods from initial RNNs \cite{amalou2022multivariate}, LSTM \cite{wang2023dafa} to recent Transformer\cite{Yuqietal2023PatchTST, zhang2023crossformer} and Graph Neural Network (GNN)\cite{Cai2024, wu2020connecting} have achieved promising performance in modeling long-range dependencies, most of these methods are confronted with a common challenge: it is hard to explicitly model the decoupling of temporal evolution patterns from variable correlations while addressing dynamic changes in data distribution simultaneously. This limitation severely hinders the forecasting performance, which further impacts the efficiency and reliability of upper-layer collaborative applications, such as scheduling among multiple agents and cross-departmental resource allocation.

Then, to solve the above problems and provide a more reliable data-driven support for collaborative design of complex systems, we aim at accurately learning the channel-temporal dependencies from complicated historical data and alleviating non-stationarity-induced generalization limitation, specifically, we develop a parallel dual-branched architecture with a linear temporal modeling layer and a channel attention mechanism to explicitly decouple and jointly learn temporal evolution patterns and multivariate interaction dependencies. Furthermore, motivated by the unavoidable common distribution shift for industrial data in distributed environment, we propose a Frequency-Domain Stationarity Correction strategy inspired by Ma et al. \cite{ma2024u} to adaptively adjust the predicted results based on the spectral characteristics and avoid model instability caused by dynamic changes. Our experiments on multiple benchmarks demonstrate that our method achieves superior performance compared with state-of-the-art models in terms of forecasting accuracy and more importantly can be a powerful engine for industrial collaborative systems.

The main innovations are shown as follows.
\begin{itemize}
\item Proposed a Dual-Branch Channel-Temporal structure to capture local temporal dynamics and variable dependencies, thereby enhancing the internal coupling representation of complex systems.
\item Proposed a Global Inter-Patch Attention Fusion mechanism to transcend local limitations and improve the learning of long-period patterns.
\item Introduced a Frequency-Domain Distribution Adaptation approach to alleviate non-stationarity and reduce the influence of distribution shift.
\end{itemize}

\section{Related Work}

\subsection{Deep Learning Models for Time Series Forecasting}
The research lines in time series forecasting have evolved from early statistical models and Recurrent Neural Networks (RNNs, LSTMs) to deep models mainly composed of Transformers and MLPs. Due to their ability to capture long-range dependencies from self-attention, Transformer-based models like Informer \cite{zhou2021informer} and Autoformer \cite{wu2021autoformer} have shown to be powerful in modeling inter-variable couplings. However, they are usually confronted with high complexity and sensitive to noises. Recently, linear and MLP-based models represented by DLinear \cite{Zeng2023} and TimeMixer \cite{wang2024timemixer} have achieved counter-intuitively better performance with simple designs, which motivates us to reconsider the necessity of using complicated Transformer architectures. Meanwhile, Graph Neural Networks (GNNs) \cite{wu2020connecting, Cai2024, wang2025sdgf} and Channel Independence methods are widely explored in modeling spatial dependencies of multivariate time series. However, existing methods usually fail to explicitly decouple local temporal evolution from global variable interactions while balancing between efficiency and expressiveness, which limit the application of our method in complex industrial collaborative scenes.

\subsection{Non-Stationarity Adaptation and Distribution Correction}
Real-world industrial data usually possess high non-stationarity, which means dynamic distribution shifts over time will badly affect model generalization. Traditional methods usually just depend on some preprocessing methods, such as differencing or normalization (e.g., RevIN)\cite{kim2021reversible}, to eliminate trends and seasonality to stabilize input distributions. However, when eliminating trends, normalization may bring more severe problems: too much normalization will also remove non-stationary features that contain important trend information, which will affect the model to learn specific temporal dependencies. Although some recent methods try to restore original distribution information in stationarized attention mechanism or denormalization \cite{liu2022non} , effective methods that can correct distribution shifts while preserving key information in both frequency and time domain still lack. Especially, using frequency-domain characteristics for adaptive stationarity correction is less explored \cite{ma2024u}.

\begin{figure*}[htb]
  \centering
  \includegraphics[width=16cm, keepaspectratio]{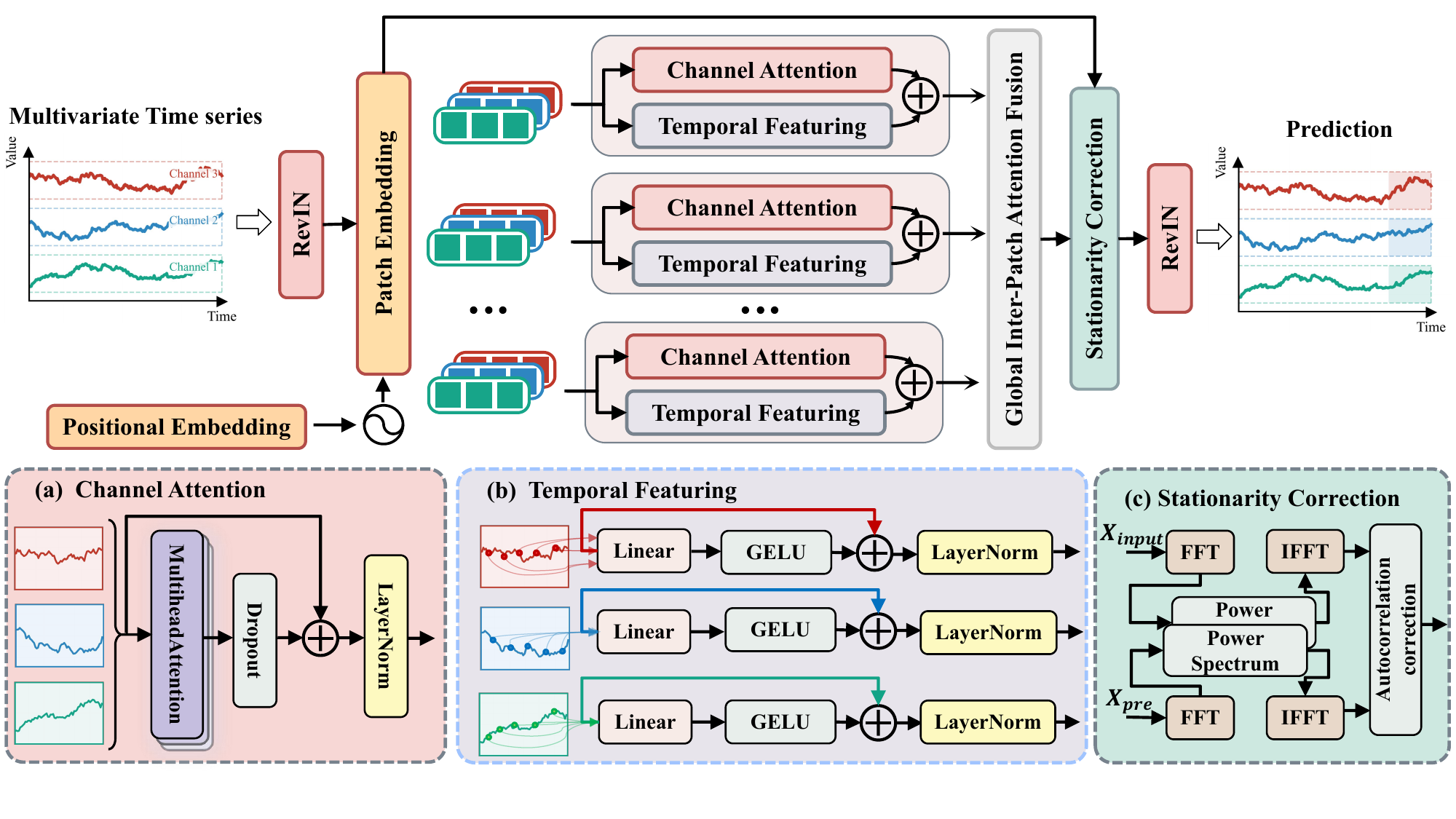}
  \vspace{-0.6cm}
  \caption{Overall workflow of the proposed D-CTNet. It integrates four stages to model channel-temporal dependencies and mitigate non-stationarity. (1) Patch Embedding and Representation Learning. (2) Dual-Branch Channel-Temporal Modeling: Channel Attention and Temporal Featuring. (3) Global Inter-patch attention. (4) Stationarity Correction.}
  \label{fig:framework}
\end{figure*}
\section{Method}
\label{sec:majhead}
\subsection{Problem Formulation}
\label{ssec:subhead}
Suppose \(\mathbf{X}_{in} \in \mathbb{R}^{L \times N}\) is a historical multivariate time series(MTS), where \(N\) is the number of variables and \(L\) is the length of historical window. The goal of the MTS task is to predict the future sequence \(\hat{\mathbf{Y}}_{\text{out}} \in \mathbb{R}^{T \times N}\) accurately in \(T\) prediction steps based on historical sequence \(\mathbf{X}_{in}\).

\subsection{Overall Framework}
As shown in Figure~\ref{fig:framework}, the overall workflow of proposed \textbf{D-CTNet} aims at efficiently modeling channel-temporal dependencies and mitigating non-stationarity in multivariate time series through four stages in an integrated way. Specifically, Patch-Based Representation Learning normalizes input via Reversible Instance Normalization (RevIN) \cite{kim2021reversible} and segments it into patches projected into a latent space with positional encodings to maintain local semantics and alleviate redundancy. Then a Dual-Branch Channel-Temporal Modeling module explicitly decouples dynamics: a linear temporal branch models intra-patch evolution, while a parallel channel attention branch captures inter-variable correlations and their features are adaptively fused. Furthermore, to overcome the limitation of local patches, Global Inter-Patch Attention Fusion employs global self-attention to facilitate cross-patch interaction and model long-range dependencies across whole history. Finally, a Frequency-Domain Stationarity Correction module adaptively corrects prediction features based on spectral autocorrelation to alleviate distribution shift and a MLP projection layer is employed to output final forecast.

\subsection{Patch Embedding and Representation Learning}
Firstly, Reversible Instance Normalization (RevIN) \cite{kim2021reversible} is implemented on input tensor \(\mathbf{X}_{in} \in \mathbb{R}^{B \times L \times C}\) , where \(B\) is batch size, \(L\) is sequence length and \(C\) is number of variables, and the normalized sequence \(\mathbf{X}_{\text{norm}} = \text{RevIN}(\mathbf{X}, \text{mode} = \text{'norm'})\) is obtained. RevIN firstly normalizes input by subtracting mean and dividing standard deviation of each instance, and then introduces learnable affine parameters ($\gamma$, $\beta$) to make the whole normalization process reversible, which further enables the prediction output to be mapped back to the original scale during inference.

Subsequently, the point-wise time series are reshaped into sequences of patches through patch embedding layer. Specifically, given patch length of P and stride of S, we cut the input sequence into N non-overlapping/partially overlapping patches. The number of patches N can be calculated as $N = \lfloor (L - P) / S \rfloor + 1$. When the external periodicity module is enabled, the patch length P can be adjusted adaptively according to periodic information of input data to improve the semantic relevance of patches. Each patch is further linearly projected to a predefined latent dimension D through 1D convolution and then added with learnable positional embeddings $\mathbf{E}_{\text{pos}}$ to preserve the temporal orderness of patches. Finally, we can obtain the tensor $\mathbf{X}_{\text{patch}} \in \mathbb{R}^{B \times C \times N \times D}$ as the input of subsequent modeling modules.

\subsection{Dual-Branch Channel-Temporal Modeling}
To simultaneously capture temporal dynamic features and multivariate correlations, we design a parallel dual-branch structure to model univariate temporal evolution patterns and inter-variables dependencies separately, and meanwhile, the efficient integration of features from two branches is achieved by implicit fusion.

\subsubsection{Linear Temporal Modeling Branch}
This branch is designed to model the temporal evolution of each individual variable. Treating patch sequence dimension N as temporal dimension, we firstly model global correlations of patch features $\mathbf{X}_{\text{patch}}$ along temporal dimension with a linear transformation $\mathbf{W}_{\text{time}}$. Subsequently, a GELU activation function is introduced to enhance the non--linear expressive capacity of model, which helps the model to fit more complex temporal patterns. Moreover, residual connection is introduced to alleviate the vanishing gradient problem during training of deep networks and preserve original feature information. Finally, layer normalization is applied to stabilize the feature distribution and training process of the model. The mathematical expression is shown as Eq. \eqref{eq:time}.
\begin{equation}\label{eq:time}
\mathbf{H}_{\text{time}} = \text{Norm}\left( \text{GELU}\left( \mathbf{X}_{\text{patch}} \mathbf{W}_{\text{time}} \right) + \mathbf{X}_{\text{patch}} \right)
\end{equation}

This module can effectively extract the global trends and periodic patterns of individual variables from neighboring time patches and provide reliable information of temporal dimension for subsequent feature fusion.

\subsubsection{Channel Attention Modeling Branch}
Parallel to the temporal modeling branch, we also attempt to model the complex dependencies between variables. We regard $C$ as the sequence dimension and apply MHA to adaptively compute the correlation weights between different variables, which enables interaction and modeling of dependencies between different variables. Then, Dropout is applied on the attention output to suppress over-fitting. After that, residual connection is used to fuse the original patch features and preserve basic information. Finally, layer normalization further improves the quality of features. The mathematical expression is shown in Eq. \eqref{eq:channel}:
\begin{equation}\label{eq:channel}
\mathbf{H}_{\text{channel}} = \text{Norm}\left( \text{Dropout}\left( \text{MHA}(\mathbf{X}_{\text{patch}}) \right) + \mathbf{X}_{\text{patch}} \right)
\end{equation}

Feature fusion of dual branches is implicitly achieved by residual addition in the channel attention branch: univariate temporal features explored by the temporal modeling branch and multivariate dependency features explored by the channel attention branch are integrated to generate a global feature representation $\mathbf{H}_{\text{fused}}$ with both temporal dynamics and variable correlations.

\subsection{Global Inter-Patch Attention Fusion}
To model long-range dependencies between patches in the whole historical window, we further apply global self-attention on the patch dimension $N$. Taking the fused features $\mathbf{H}^{\text{fused}}$ from the dual branches as input, this mechanism enables each patch to globally attend to all other patches in the whole historical window via Multi-Head Attention on the time patch axis ($\text{MHA}_{\text{patch}}$). It is realized by residual addition to enable cross-patch information interaction and global dependency modeling. Then, Dropout is applied on the attention output to suppress over-fitting, residual connection is used to fuse the original fused features and preserve key information, and finally layer normalization further improves the feature quality. The mathematical expression is shown in Eq. \eqref{eq:global}:
\begin{equation}\label{eq:global}
\mathbf{H}_{\text{global}} = \text{Norm}\left( \text{Dropout}\left( \text{MHA}_{\text{patch}}(\mathbf{H}_{\text{fused}}) \right) + \mathbf{H}_{\text{fused}} \right)
\end{equation}

The goal of this module is to break the information limitation of local patches and model long-range temporal dependencies at a global scale. In addition, this module also provides feature support from a global perspective for the final prediction.

\subsection{Frequency-Domain Stationarity Correction}

To mitigate the impact of non-stationarity of time series on prediction performance, we propose a Frequency-Domain Stationarity Correction module (S-Correction) that adaptively corrects the predicted features based on the frequency-domain autocorrelation characteristics of input and prediction features, making the frequency distribution of predicted features consistent with that of input features and thus improving the generalization ability of the model.
Denote $\mathbf{H}_{\text{global}}$ as the prediction feature output of global attention module and $\mathbf{X}_{\text{patch}}$ as the original patch input feature. The correction process is shown as follows.
Firstly, an orthogonal normalized FFT is applied on both the prediction input feature and input feature to map the time-domain features into frequency domain, i.e.

\begin{equation}
\mathbf{F}_{\text{pred}} = \text{FFT}(\mathbf{H}_{\text{global}}), \quad
\mathbf{F}_{\text{input}} = \text{FFT}(\mathbf{X}_{\text{patch}})
\end{equation}
Then, the power spectrum is computed as the dot product of complex valued signal and its complex conjugate, which describes the energy distribution of the signal in different frequencies.
\begin{equation}
S_{\text{pred}} = \mathbf{F}_{\text{pred}} \cdot \mathbf{F}_{\text{pred}}^*, \quad
S_{\text{input}} = \mathbf{F}_{\text{input}} \cdot \mathbf{F}_{\text{input}}^*
\end{equation}
Next, the IFFT is applied on the power spectrum to map them back to time domain. The clamp operation is also applied to avoid interference from negative energy in correlation computations.
\begin{equation}
\begin{aligned}
\hat{S}_{\text{pred}} &= \text{clamp}(\text{IFFT}(S_{\text{pred}}), 0, \infty),\\
\hat{S}_{\text{input}} &= \text{clamp}(\text{IFFT}(S_{\text{input}}), 0, \infty)
\end{aligned}
\end{equation}
Finally, the correction factor $\alpha$ is computed as the cross-correlation between predicted input and input autocorrelation.
\begin{equation}
\alpha = \sqrt{\frac{\sum \hat{S}_{\text{pred}} \cdot \hat{S}_{\text{input}}}{\sum \hat{S}_{\text{input}}^2 + \epsilon}}
\end{equation}

where $\epsilon$ is a small value to avoid division by zero.
Lastly, the prediction features $\mathbf{H}_{\text{global}}$ are element-wise multiplied by $\alpha$ and:
\begin{equation}
\mathbf{H}_{\text{final}} = \mathbf{H}_{\text{global}} \cdot \alpha
\end{equation}
Finally the flattened prediction features are projected to the target prediction dimension using a linear layer and passed through a RevIN denormalization layer to rescale the output to the input scale, resulting in final prediction $\hat{\mathbf{Y}}_{\text{out}}$.

\begin{table*}[htbp]
  \centering
  \caption{Quantitative evaluation of time series forecasting models including Transformer, GNN, MLP, CNN, and RNN architectures.}
  \resizebox{\textwidth}{!}{%
    \setlength{\tabcolsep}{2.5pt} 
    \renewcommand{\arraystretch}{0.9} 
    \begin{tabular}{cc|cccccccccccccccccc}
    \toprule
    \multicolumn{2}{c|}{\multirow{2}{*}{\textbf{Models}}} & \multicolumn{2}{c}{\multirow{2}{*}{\textbf{Ours}}} & \multicolumn{4}{c}{Transformer} & \multicolumn{4}{c}{GNN} & \multicolumn{4}{c}{MLP} & \multicolumn{2}{c}{CNN} & \multicolumn{2}{c}{RNN} \\
    \cmidrule(lr){5-8} \cmidrule(lr){9-12} \cmidrule(lr){13-16} \cmidrule(lr){17-18} \cmidrule(lr){19-20}
    \multicolumn{2}{c|}{} & \multicolumn{2}{c}{} & \multicolumn{2}{c}{Crossformer} & \multicolumn{2}{c}{PatchTST} & \multicolumn{2}{c}{MSGNet} & \multicolumn{2}{c}{MTGNN} & \multicolumn{2}{c}{RLinear} & \multicolumn{2}{c}{DLinear} & \multicolumn{2}{c}{TimesNet} & \multicolumn{2}{c}{LSSL} \\
    \midrule
    \multicolumn{2}{c|}{Metric} & MSE & MAE & MSE & MAE & MSE & MAE & MSE & MAE & MSE & MAE & MSE & MAE & MSE & MAE & MSE & MAE & MSE & MAE \\
    \midrule
    \multicolumn{1}{c|}{\multirow{5}[2]{*}{\begin{sideways}ETTm1\end{sideways}}} 
    & 96  & \underline{0.337} & \textbf{0.365} & 0.404 & 0.426 & 0.352 & 0.374 & \textbf{0.319} & \underline{0.366} & 0.381 & 0.415 & 0.355 & 0.376 & 0.345 & 0.372 & 0.338 & 0.375 & 0.450 & 0.477 \\
    \multicolumn{1}{c|}{} 
    & 192 & \textbf{0.376} & \textbf{0.387} & 0.450 & 0.451 & 0.374 & \underline{0.387} & \underline{0.376} & 0.397 & 0.442 & 0.451 & 0.391 & 0.392 & 0.380 & 0.389 & 0.374 & 0.387 & 0.469 & 0.481 \\
    \multicolumn{1}{c|}{} 
    & 336 & \textbf{0.408} & \textbf{0.411} & 0.532 & 0.515 & 0.421 & 0.414 & 0.417 & 0.422 & 0.475 & 0.475 & 0.424 & 0.415 & 0.413 & 0.413 & \underline{0.410} & \underline{0.411} & 0.583 & 0.574 \\
    \multicolumn{1}{c|}{} 
    & 720 & 0.470 & \textbf{0.445} & 0.666 & 0.589 & \textbf{0.462} & \underline{0.449} & 0.481 & 0.458 & 0.531 & 0.507 & 0.487 & 0.450 & 0.474 & 0.453 & 0.478 & 0.450 & 0.632 & 0.596 \\
    \multicolumn{1}{c|}{} 
    & Avg & \textbf{0.398} & \textbf{0.402} & 0.513 & 0.495 & 0.402 & \underline{0.406} & 0.398 & 0.411 & 0.457 & 0.462 & 0.414 & 0.408 & 0.403 & 0.407 & 0.400 & \underline{0.406} & 0.534 & 0.532 \\
    \midrule
    \multicolumn{1}{c|}{\multirow{5}[2]{*}{\begin{sideways}ETTm2\end{sideways}}} 
    & 96  & \textbf{0.177} & \textbf{0.255} & 0.287 & 0.366 & 0.183 & 0.270 & \underline{0.177} & \underline{0.262} & 0.240 & 0.343 & 0.182 & 0.265 & 0.193 & 0.292 & 0.187 & 0.267 & 0.243 & 0.342 \\
    \multicolumn{1}{c|}{} 
    & 192 & \textbf{0.245} & \textbf{0.303} & 0.414 & 0.492 & 0.255 & 0.314 & 0.247 & 0.307 & 0.398 & 0.454 & \underline{0.246} & \underline{0.304} & 0.284 & 0.362 & 0.249 & 0.309 & 0.392 & 0.448 \\
    \multicolumn{1}{c|}{} 
    & 336 & \textbf{0.301} & \textbf{0.341} & 0.597 & 0.542 & 0.309 & 0.347 & 0.312 & 0.346 & 0.568 & 0.550 & \underline{0.307} & \underline{0.342} & 0.369 & 0.427 & 0.321 & 0.351 & 0.932 & 0.724 \\
    \multicolumn{1}{c|}{} 
    & 720 & \underline{0.410} & \underline{0.404} & 1.730 & 1.042 & 0.412 & 0.404 & 0.414 & 0.403 & 1.072 & 0.767 & \textbf{0.407} & \textbf{0.398} & 0.554 & 0.522 & 0.408 & 0.403 & 1.372 & 0.879 \\
    \multicolumn{1}{c|}{} 
    & Avg & \textbf{0.283} & \textbf{0.326} & 0.757 & 0.611 & 0.290 & 0.334 & 0.288 & 0.330 & 0.570 & 0.529 & \underline{0.286} & \underline{0.327} & 0.350 & 0.401 & 0.291 & 0.333 & 0.735 & 0.598 \\
    \midrule
    \multicolumn{1}{c|}{\multirow{5}[2]{*}{\begin{sideways}ETTh1\end{sideways}}} 
    & 96  & \textbf{0.340} & \textbf{0.375} & 0.423 & 0.448 & 0.460 & 0.447 & 0.390 & 0.411 & 0.440 & 0.450 & 0.386 & \underline{0.395} & 0.386 & 0.401 & \underline{0.384} & 0.402 & 0.548 & 0.528 \\
    \multicolumn{1}{c|}{} 
    & 192 & \textbf{0.341} & \textbf{0.380} & 0.471 & 0.474 & 0.477 & 0.429 & 0.442 & 0.442 & 0.449 & 0.433 & 0.437 & \underline{0.424} & 0.437 & 0.432 & \underline{0.436} & 0.429 & 0.542 & 0.526 \\
    \multicolumn{1}{c|}{} 
    & 336 & \textbf{0.342} & \textbf{0.388} & 0.570 & 0.546 & 0.546 & 0.496 & 0.480 & 0.468 & 0.598 & 0.554 & \underline{0.479} & \underline{0.446} & 0.481 & 0.459 & 0.491 & 0.469 & 1.298 & 0.942 \\
    \multicolumn{1}{c|}{} 
    & 720 & \textbf{0.412} & \textbf{0.453} & 0.653 & 0.621 & 0.544 & 0.517 & 0.494 & 0.488 & 0.685 & 0.620 & \underline{0.481} & \underline{0.470} & 0.519 & 0.516 & 0.521 & 0.501 & 0.721 & 0.659 \\
    \multicolumn{1}{c|}{} 
    & Avg & \textbf{0.359} & \textbf{0.399} & 0.529 & 0.522 & 0.507 & 0.472 & 0.452 & 0.452 & 0.543 & 0.514 & \underline{0.446} & \underline{0.434} & 0.456 & 0.452 & 0.458 & 0.450 & 0.777 & 0.664 \\
    \midrule
    \multicolumn{1}{c|}{\multirow{5}[2]{*}{\begin{sideways}ETTh2\end{sideways}}} 
    & 96  & \underline{0.306} & \underline{0.353} & 0.745 & 0.584 & 0.308 & 0.355 & 0.328 & 0.371 & 0.496 & 0.509 & \textbf{0.288} & \textbf{0.338} & 0.333 & 0.387 & 0.340 & 0.374 & 1.616 & 1.036 \\
    \multicolumn{1}{c|}{} 
    & 192 & \underline{0.389} & \underline{0.402} & 0.877 & 0.656 & 0.393 & 0.405 & 0.402 & 0.414 & 0.716 & 0.616 & \textbf{0.374} & \textbf{0.390} & 0.477 & 0.476 & 0.402 & 0.414 & 2.083 & 1.197 \\
    \multicolumn{1}{c|}{} 
    & 336 & \textbf{0.414} & \underline{0.434} & 1.043 & 0.731 & 0.427 & 0.436 & 0.435 & 0.430 & 0.718 & 0.614 & \underline{0.415} & \textbf{0.426} & 0.594 & 0.541 & 0.452 & 0.452 & 2.970 & 1.439 \\
    \multicolumn{1}{c|}{} 
    & 720 & \underline{0.435} & \underline{0.450} & 1.104 & 0.763 & 0.436 & 0.450 & 0.417 & 0.441 & 1.161 & 0.791 & \textbf{0.420} & \textbf{0.440} & 0.831 & 0.657 & 0.462 & 0.468 & 2.576 & 1.363 \\
    \multicolumn{1}{c|}{} 
    & Avg & 0.386 & \underline{0.410} & 0.942 & 0.684 & 0.391 & 0.412 & 0.396 & 0.414 & 0.773 & 0.633 & \textbf{0.374} & \textbf{0.399} & 0.559 & 0.515 & 0.414 & 0.427 & 2.311 & 1.259 \\
    \midrule
    \multicolumn{1}{c|}{\multirow{5}[2]{*}{\begin{sideways}Electricity\end{sideways}}} 
    & 96  & \textbf{0.152} & \textbf{0.246} & 0.219 & 0.314 & 0.190 & 0.296 & \underline{0.165} & \underline{0.274} & 0.211 & 0.305 & 0.201 & 0.281 & 0.197 & 0.282 & 0.168 & 0.272 & 0.300 & 0.392 \\
    \multicolumn{1}{c|}{} 
    & 192 & \textbf{0.168} & \textbf{0.259} & 0.231 & 0.322 & 0.199 & 0.304 & \underline{0.184} & 0.292 & 0.225 & 0.319 & 0.201 & 0.283 & 0.196 & 0.285 & \underline{0.184} & 0.289 & 0.297 & 0.390 \\
    \multicolumn{1}{c|}{} 
    & 336 & \textbf{0.185} & \textbf{0.277} & 0.246 & 0.337 & 0.217 & 0.319 & \underline{0.195} & 0.302 & 0.247 & 0.340 & 0.215 & 0.298 & 0.209 & 0.301 & 0.198 & 0.300 & 0.317 & 0.403 \\
    \multicolumn{1}{c|}{} 
    & 720 & 0.222 & \textbf{0.308} & 0.280 & 0.363 & 0.258 & 0.352 & 0.231 & 0.332 & 0.287 & 0.373 & 0.257 & 0.331 & 0.245 & 0.333 & \textbf{0.220} & \underline{0.320} & 0.338 & 0.417 \\
    \multicolumn{1}{c|}{} 
    & Avg & \textbf{0.182} & \textbf{0.273} & 0.244 & 0.334 & 0.216 & 0.318 & 0.194 & 0.300 & 0.243 & 0.334 & 0.219 & 0.298 & 0.212 & 0.300 & \underline{0.193} & 0.295 & 0.313 & 0.401 \\
    \midrule
    \multicolumn{1}{c|}{\multirow{5}[2]{*}{\begin{sideways}Weather\end{sideways}}} 
    & 96  & \textbf{0.162} & \textbf{0.201} & 0.158 & 0.230 & 0.186 & 0.227 & \underline{0.163} & \underline{0.212} & 0.171 & 0.231 & 0.192 & 0.232 & 0.196 & 0.255 & 0.172 & 0.220 & 0.174 & 0.252 \\
    \multicolumn{1}{c|}{} 
    & 192 & \textbf{0.210} & \textbf{0.254} & 0.206 & 0.277 & 0.234 & 0.265 & \underline{0.212} & \underline{0.254} & 0.215 & 0.274 & 0.240 & 0.271 & 0.237 & 0.296 & 0.219 & 0.261 & 0.238 & 0.313 \\
    \multicolumn{1}{c|}{} 
    & 336 & 0.276 & \textbf{0.289} & 0.272 & 0.335 & 0.284 & 0.301 & 0.272 & \underline{0.299} & \textbf{0.266} & 0.313 & 0.292 & 0.307 & 0.283 & 0.335 & 0.280 & 0.306 & 0.287 & 0.355 \\
    \multicolumn{1}{c|}{} 
    & 720 & \textbf{0.341} & \underline{0.350} & 0.398 & 0.418 & 0.356 & 0.349 & 0.350 & \textbf{0.348} & \underline{0.344} & 0.375 & 0.364 & 0.353 & 0.345 & 0.381 & 0.365 & 0.359 & 0.384 & 0.415 \\
    \multicolumn{1}{c|}{} 
    & Avg & \textbf{0.247} & \textbf{0.274} & 0.259 & 0.315 & 0.265 & 0.286 & \underline{0.249} & \underline{0.278} & \underline{0.249} & 0.298 & 0.272 & 0.291 & 0.265 & 0.317 & 0.259 & 0.287 & 0.271 & 0.334 \\
    \midrule
    \multicolumn{1}{c|}{\multirow{5}[2]{*}{\begin{sideways}Exchange\end{sideways}}} 
    & 96  & \textbf{0.087} & \textbf{0.210} & 0.256 & 0.367 & \underline{0.088} & \underline{0.205} & 0.102 & 0.230 & 0.267 & 0.378 & 0.093 & 0.217 & \underline{0.088} & 0.218 & 0.107 & 0.234 & 0.395 & 0.474 \\
    \multicolumn{1}{c|}{} 
    & 192 & \textbf{0.176} & \underline{0.300} & 0.470 & 0.509 & \underline{0.176} & \textbf{0.299} & 0.195 & 0.317 & 0.590 & 0.578 & 0.184 & 0.307 & \underline{0.176} & 0.315 & 0.226 & 0.344 & 0.776 & 0.698 \\
    \multicolumn{1}{c|}{} 
    & 336 & \underline{0.311} & \textbf{0.414} & 1.268 & 0.883 & \textbf{0.301} & \underline{0.397} & 0.359 & 0.436 & 0.939 & 0.749 & 0.351 & 0.432 & 0.313 & 0.427 & 0.367 & 0.448 & 1.029 & 0.797 \\
    \multicolumn{1}{c|}{} 
    & 720 & \textbf{0.812} & \textbf{0.694} & 1.767 & 1.068 & 0.901 & 0.714 & 0.940 & 0.738 & 1.107 & 0.834 & 0.886 & 0.714 & 0.839 & 0.695 & 0.964 & 0.746 & 2.283 & 1.222 \\
    \multicolumn{1}{c|}{} 
    & Avg & \textbf{0.347} & \underline{0.405} & 0.940 & 0.707 & 0.367 & \textbf{0.404} & 0.399 & 0.430 & 0.726 & 0.635 & 0.379 & 0.418 & 0.354 & 0.414 & 0.416 & 0.443 & 1.121 & 0.798 \\
    \midrule
    \multicolumn{2}{c|}{$1^{st}$ Count} & \multicolumn{2}{c|}{51} & \multicolumn{2}{c|}{0} & \multicolumn{2}{c|}{4} & \multicolumn{2}{c|}{2} & \multicolumn{2}{c|}{1} & \multicolumn{2}{c|}{11} & \multicolumn{2}{c|}{0} & \multicolumn{2}{c|}{1} & \multicolumn{2}{c}{0} \\
    \bottomrule
    \end{tabular}%
  }
  \label{tab:results}%
\end{table*}

\section{EXPERIMENTS}
\label{sec:EXPERIMENTS}
\subsection{Experimental Setup}\label{ssec:subhead}

\noindent\textbf{Datasets.} We conduct experiments on seven popular time-series benchmarks: Electricity Transformer Temperature datasets (ETTh1, ETTh2, ETTm1, ETTm2), Exchange-Rate, Electricity, Traffic and Weather. All the datasets are split in a chronological way to maintain the temporal order. For ETT series, we follow \cite{wu2020connecting} to split the dataset into training, validation and testing sets with a ratio of 6:2:2, which is an appropriate way to balance the amount of historical data used and the generalization evaluated. For other datasets we follow \cite{wu2023timesnet, Zeng2023} to split them into training, validation and testing sets with a ratio of 7:1:2.

\noindent\textbf{Baselines.} We compare our method with eight state-of-the-art methods from different paradigms: GNN based MSGNet \cite{Cai2024} and MTGNN \cite{wu2020connecting}, CNN based TimesNet \cite{wu2023timesnet}, Transformer based PatchTST \cite{Yuqietal2023PatchTST} and Crossformer \cite{zhang2023crossformer}, MLP based DLinear \cite{Zeng2023} and RLinear \cite{toner2024analysis}, and RNN based LSSL \cite{gu2021combining}. These methods cover various kinds of architecture, which can serve as a comprehensive benchmark to evaluate the performance of our method.

\noindent\textbf{Implementation.} All the experiments are conducted in PyTorch on NVIDIA RTX 4090 GPU (24 GB). We train the model with Adam optimizer with batchsize 32. Following the long-term forecasting protocols in \cite{wu2020connecting}, input sequence length is $L=96$, and we evaluate the prediction horizon of $T \in \{96, 192, 336, 720\}$ time steps. The evaluation metric is Mean Squared Error (MSE) and Mean Absolute Error (MAE) \cite{Cai2024}.

\subsection{Results and Analysis}\label{ssec:subhead}

\begin{figure}[h]
\setlength{\abovecaptionskip}{2pt} 
\setlength{\belowcaptionskip}{2pt} 

\begin{minipage}[b]{0.49\linewidth}
  \centering
  \includegraphics[width=3.9cm]{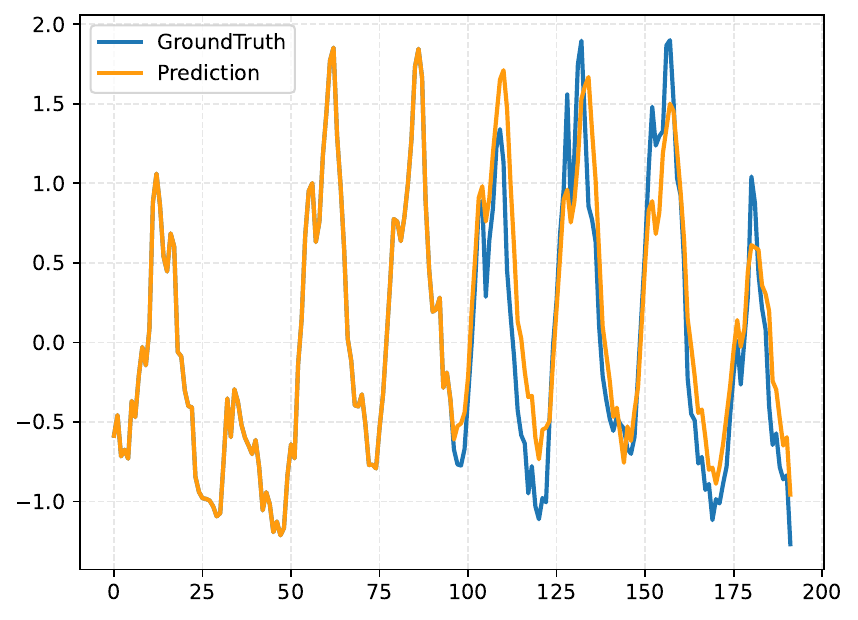}
  \centerline{(a) Electricity, output-96}
\end{minipage}
\hfill
\begin{minipage}[b]{0.49\linewidth}
  \centering
  \includegraphics[width=3.9cm]{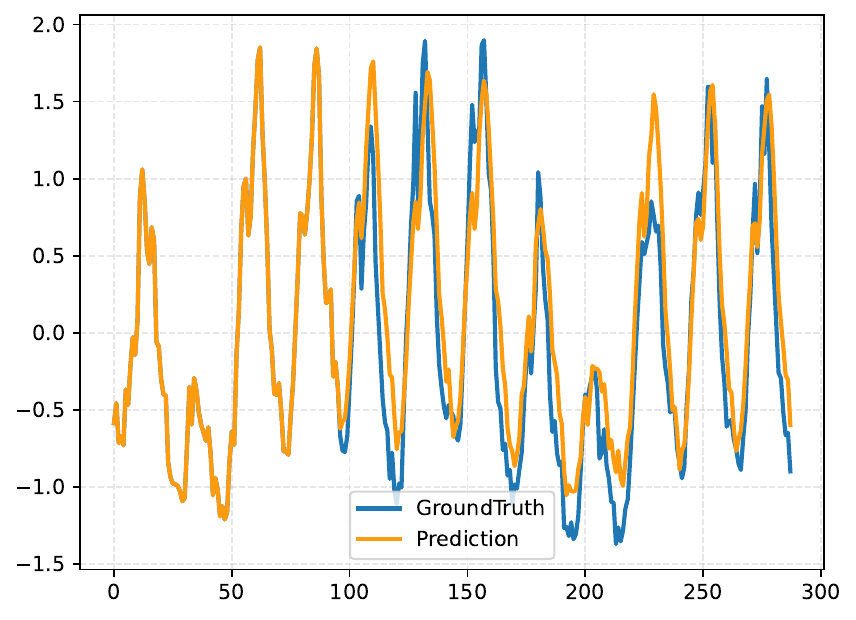}
  \centerline{(b) Electricity, output-192}
\end{minipage}

\begin{minipage}[b]{0.49\linewidth}
  \centering
  \includegraphics[width=3.9cm]{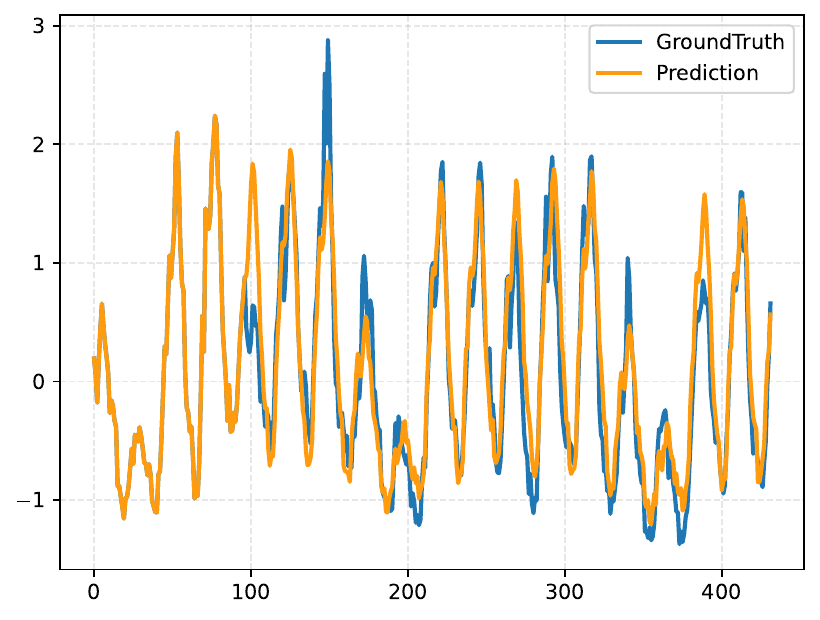}
  \centerline{(c) Electricity, output-336}
\end{minipage}
\hfill
\begin{minipage}[b]{0.49\linewidth}
  \centering
  \includegraphics[width=3.9cm]{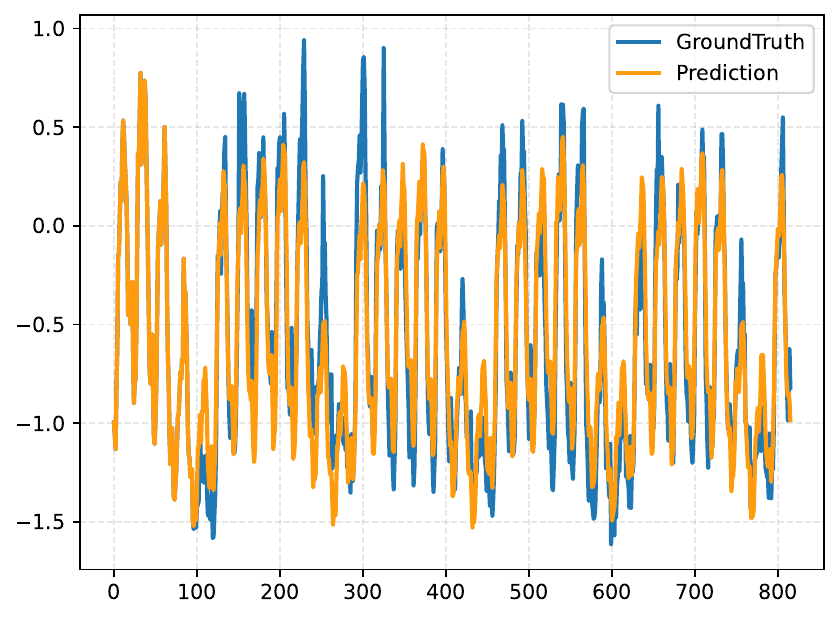}
  \centerline{(d) Electricity, output-720}
\end{minipage}

\caption{Visualization comparison of forecasting results.}
\label{fig:res}
\end{figure}

As shown in Table \ref {tab:results}, our model obtains the best performance in most experimental scenarios, and the best performance in most benchmark testing methods. Especially on ETTm2, ETTh1 and Electricity datasets, our model is the first place in most MSE/MAE performance of different step size. Our model is much more robust to non-stationary data compared with GNN-based method (MSGNet) and Transformer-based method (PatchTST). These methods are usually depended on fixed structure or pure attention mechanism. However, D-CTNet is modeled by dual-branch architecture and frequency-domain correction module, which can dynamically describe the time-varying correlations between variables and data distribution change. Figure~\ref{fig:res} presents the visualization of prediction results on the Electricity dataset. In addition, the advantages of our model in long prediction step size (e.g., 720-step in Exchange) also proves the effectiveness of global patch fusion module in modeling long-range dependencies.

\begin{table}[htbp]
\centering
\caption{Ablation analysis of D-CTNet.}
\label{tab:my_label}
\renewcommand{\arraystretch}{1.05}  
\resizebox{\columnwidth}{!}{
\begin{tabular}{ccccccc}
\hline
Dataset & \multicolumn{2}{c}{ETTh1} & \multicolumn{2}{c}{Exchange} & \multicolumn{2}{c}{Weather} \\ \hline
Metric  & MSE     & MAE      & MSE     & MAE      & MSE      & MAE      \\ \hline
D-CTNet  & \textbf{0.340} & \textbf{0.375}& \textbf{0.087} & \textbf{0.210}& \textbf{0.162} & \textbf{0.201}    \\
w/o-DBCT     & 0.401      & 0.427    & 0.152         & 0.319       & 0.213       & 0.260    \\
w/o-GPAF      & 0.386      & 0.394     & 0.112       & 0.270     & 0.185             & 0.241     \\
w/o-FSC     & 0.361      & 0.411    & 0.098         & 0.220     & 0.174          & 0.230    \\
\hline
\end{tabular}}
\end{table}

\subsection{Ablation Study}
\label{ssec:subhead}

Table~\ref{tab:my_label} shows the ablation studies on ETTh1, Exchange, and Weather datasets. \textbf{w/o-DBCT} means that we remove the Dual-Branch Channel-Temporal module. From the results, we can make the following observations: (1) Compared with all other models, removing the Dual-Branch Channel-Temporal module (w/o-DBCT) incurs the largest performance drops on these three benchmarks. Especially on the Exchange dataset, the MSE increases from 0.087 to 0.152. It shows that explicitly decoupling temporal evolution and multivariate correlation is beneficial for modeling the dynamics of industrial collaborative data. (2) Removing the Global Inter-Patch Attention Fusion (w/o-GPAF) leads to a relatively large performance drop. Especially on the Weather dataset, the MSE increases to 0.185. It shows that global inter-patch attention fusion is important for modeling long-range temporal dependency of meteorological data. (3) When disabling the Frequency-Domain Stationarity Correction (w/o-FSC), the MSE increases. It shows that frequency sub-bands with spectral alignment can reduce the generalization limitation of non-stationarity on target time series.

\section{CONCLUSION}
\label{sec:foot}
In this paper, we propose Patch-Based Dual-Branch Channel-Temporal Forecasting Network for collaborative sensing in complex industrial systems. We explicitly decouple the temporal evolution and multivariate dependent modeling by designing a parallel architecture of linear temporal modeling and channel attention. Besides, the frequency-domain stationarity correction mechanism can alleviate the generalization limitation caused by data non-stationarity. The experimental results show that our model achieves higher accuracy and robustness than other state-of-the-art approaches. Our method provides reliable data-driven support for Digital Twin construction and Industrial Internet of Things collaborative decision-making. In the future, we will explore lightweight model deployment on edge devices and cross-domain integration with Large Language Models.

\section{Acknowledgment}
This work was supported by the National Natural Science Foundation of China (Grant No.62372366).

\bibliographystyle{IEEEtran}
\bibliography{ref}

\end{document}